\documentclass[10pt,twocolumn,letterpaper]{article}

\usepackage{cvpr}              % To produce the CAMERA-READY version
\usepackage{amsmath,amsfonts,bm}

\def\eqref#1{equation~\ref{#1}}
\def\1{\bm{1}}

\DeclareMathAlphabet{\mathsfit}{\encodingdefault}{\sfdefault}{m}{sl}
\SetMathAlphabet{\mathsfit}{bold}{\encodingdefault}{\sfdefault}{bx}{n}

\usepackage{multirow}
\usepackage{url}
\usepackage{graphicx}
\usepackage{booktabs}
\usepackage{colortbl}
\usepackage{xcolor}
\usepackage{xspace}
\usepackage{wrapfig}
\usepackage{caption}
\usepackage{amsmath}
\usepackage{adjustbox}
\usepackage{longtable}
\usepackage{algorithm}
\usepackage{algorithmic}

\newcommand{\sysName}{\textsc{PTA}\xspace}
\newcommand{\fei}[1]{#1}

\usepackage[accsupp]{axessibility} 

\definecolor{cvprblue}{rgb}{0.21,0.49,0.74}
\definecolor{mypink}{rgb}{1,0.4,0.6}
\usepackage[pagebackref,breaklinks,colorlinks,allcolors=cvprblue,urlcolor=mypink]{hyperref}
\def\paperID{5} % *** Enter the Paper ID here
\def\confName{CVPR}
\def\confYear{2026}

\title{Purify-then-Align: Towards Robust Human Sensing under Modality Missing with Knowledge Distillation from Noisy Multimodal Teacher}

\author{Pengcheng Weng$^{1,2*}$, Yanyu Qian$^{3,1*}$, Yangxin Xu$^{1*}$, Fei Wang$^{1\dagger}$ \\
\textit{$^1$ School of Software Engineering, Xi'an Jiaotong University, China}\\
 \textit{$^2$ Institute of Computer Science, Universität Bern, Switzerland} \\
\textit{$^3$ College of Computing and Data Science, Nanyang Technological University, Singapore}\\
{\tt \small  pengcheng.weng@students.unibe.ch, yanyu003@e.ntu.edu.sg}\\ {\tt \small  li7032202@stu.xjtu.edu.cn,  feynmanw@xjtu.edu.cn}\\
{\small $*$Equal contribution, $\dagger$Corresponding author and project lead}\\
{\small \url{https://github.com/Vongolia11/PTA}}
}
\begin{document}
\maketitle

%%%%%%%%% ABSTRACT
\begin{abstract}

Robust multimodal human sensing must overcome the critical challenge of missing modalities.
Two principal barriers are the Representation Gap between heterogeneous data and the Contamination Effect from low-quality modalities. 
These barriers are causally linked as the corruption introduced by contamination fundamentally impedes the reduction of representation disparities.
In this paper, we propose \sysName, a novel ``Purify-then-Align" framework that solves this causal dependency through a synergistic integration of meta-learning and knowledge diffusion. 
To purify the knowledge source, \sysName first employs a meta-learning-driven weighting mechanism that dynamically learns to down-weight the influence of noisy, low-contributing modalities. Subsequently, to align different modalities, 
\sysName introduces a diffusion-based knowledge distillation paradigm where an information-rich clean teacher, formed from this purified consensus, refines the features of each student modality. The ultimate payoff of this ``Purify-then-Align" strategy is the creation of exceptionally powerful single-modality encoders, imbued with cross-modal knowledge. Comprehensive experiments on the large-scale MM-Fi and XRF55 datasets, under pronounced Representation Gap and Contamination Effect, demonstrate that \sysName achieves state-of-the-art performance, significantly improving the robustness of single-modality models in diverse missing-modality scenarios.

\end{abstract}

\section{Introduction}\label{sec:introduction}

Multimodal human sensing, the art of capturing and interpreting human activities and states through various sensors, stands as a cornerstone technology in frontier AI domains such as human-computer interaction, intelligent healthcare, and the metaverse. The fusion of multiple modalities, for instance, the dense information from depth cameras, the precise 3D structure from LiDAR, and the all-weather perceptual capabilities of wearable devices, offers a path to surmount the limitations of any single sensor, promising systems with superior accuracy and robustness~\citep{baltruvsaitis2018multimodal}. 
However, in real-world scenarios, the issue of missing modalities due to hardware failures, environmental interference (e.g., adverse weather for LiDAR), deployment cost constraints, or communication dropouts~\citep{ma2021smil}, represents a pervasive challenge that severely degrades the performance of multimodal fusion models~\cite{wu2024deep}.

The multimodal human sensing problem is further compounded by two fundamental challenges: The first is the Representation Gap. The physical principles of different sensors are profoundly dissimilar, leading to a vast chasm between their data representations (e.g., the grid-like pixels of an image versus the sparse point cloud from LiDAR)~\citep{li2022deepfusion}. Directly combining these heterogeneous features often results in the loss or misinterpretation of critical information. The second is the Contamination Effect. The signal-to-noise ratio and task-specific contribution of modalities can vary dramatically. When a high-quality modality is fused with a low-quality, high-noise one, the uncertainty from the latter inevitably contaminates the former, degrading overall performance. To address these issues, the research community has explored several avenues. Generative approaches, such as VAEs~\citep{nazabal2020handling} or GANs~\citep{wang2023incomplete}, attempt to reconstruct missing features but are often plagued by training instability, high computational costs, and a tendency to hallucinate inaccurate details. Shared representation learning~\citep{xu2024leveraging, wang2020icmsc, liu2021incomplete} aims to project all modalities into a common latent space, but often struggles to preserve unique critical information from highly heterogeneous modalities~\citep{liang2022foundations}. Simpler fusion mechanisms~\citep{parthasarathy2020training,cheng2024novel,yang2024incomplete} like averaging are highly susceptible to the Contamination Effect, while traditional knowledge distillation~\citep{xue2022modality,kong2019mmact,wang2020incomplete,kwak2025cross} fails to bridge the immense Representation Gap in multimodal contexts~\citep{xue2022modality, liu2024contrastive}. In summary, while existing methods have made progress on individual aspects, they often fail to address the underlying connection between the Representation Gap and the Contamination Effect. A truly unified framework that understands and resolves these two challenges remains an open challenge.

A promising approach to robust human sensing under modal absence is to enable each unimodal student to operate independently by learning from an all-modality teacher consensus. However, such a teacher-student paradigm faces two intertwined challenges. First, significant Representation Gaps among modalities in the teacher can impede effective knowledge transfer to the student. Second, due to the Contamination Effect, the teacher's representation is often contaminated by low-quality or noisy modalities. These issues are not merely correlated; they are causally linked. Therefore, building a robust framework necessitates ensuring the purity of the knowledge source at the outset. In this paper, we propose \sysName, a novel framework built on a synergistic ``Purify-then-Align" paradigm.

\begin{itemize}
    \item Purify (Sec~\ref{sec:purifying}): To solve the Contamination Effect, \sysName first purifies the teacher. It employs a meta-learning-driven weighting mechanism that dynamically learns to identify and down-weight the influence of noisy, low-contributing modalities during fusion. This ensures the creation of a clean, high-quality consensus teacher.
    \item Align (Sec~\ref{sec:aligning}): Then, \sysName proceeds to align the students to bridge the Representation Gap with a diffusion-based knowledge distillation paradigm that empowers this clean teacher to refine the features of each student modality.
\end{itemize}
The ultimate payoff of this ``Purify-then-Align" strategy is the creation of exceptionally powerful single-modality encoders. By learning from a clean, all-knowing teacher, each student model is imbued with rich, cross-modal knowledge, making it the key to achieving true robustness when other sensors fail. 
The contributions of this work are as follows:
\begin{itemize}
        \item We propose ``Purify-then-Align", a novel framework that, for the first time, identifies and solves the causal link between the fundamental issues of Contamination Effect and the Representation Gap. 
        \item We design a meta-learning weighting mechanism to ``Purify" the multimodal teacher, effectively mitigating the negative impact of low-quality modalities during fusion.
        \item We introduce a knowledge diffusion-based alignment strategy that uses this clean teacher to significantly enhance the capability of each single-modality feature encoder, thereby bridging the Representation Gap. 
        \item We conducted comprehensive experiments on the large-scale multimodal dataset MM-Fi~\citep{yang2023mm} and XRF55~\citep{lan2025xrf}. Our results show that \sysName achieves state-of-the-art performance in numerous combinations of missing modality. 
\end{itemize}

\section{Related Work}\label{sec:related-work}

\subsection{Human Sensing}\label{sec:human-sensing}
Human sensing is a key research area that uses multi-source sensors to recognize human behavior, posture, and physiological states. In vision-based sensing, early foundational works like OpenPose~\citep{cao2019openpose} enabled real-time 2D multi-person keypoint detection, while subsequent models like PoseConv3D~\citep{duan2022revisiting} advanced the field by capturing spatio-temporal dynamics for 3D pose estimation directly from video. As a non-intrusive alternative, radio frequency sensing has also seen significant progress. Millimeter-wave signals have been used to reconstruct fine-grained 3D body meshes~\citep{xue2021mmmesh,ding2023mi}, for gait-based human recognition~\citep{ozturk2021gaitcube}, hand poses~\cite{lv2025mmegohand}, and to estimate scene flow to enhance human activity recognition and body part tracking~\citep{ding2024milliflow}. Meanwhile, Wi-Fi sensing has been applied to passive activity recognition~\citep{zhang2022wi,wang2019joint} and human pose estimation~\citep{wang2019person,yan2024person}. other researchers have designed custom radar systems for pose estimation~\citep{zhao2018through} and mesh reconstruction~\citep{zhao2019through}.

To overcome the weaknesses of single modalities, such as vision's sensitivity to occlusion, the dominant trend is to fuse multimodal data for enhanced robustness and accuracy. This paradigm has led to works combining Wi-Fi, mmWave radar, camera, lidar, and RFID signals~\citep{yang2023mm, wang2024xrf55} for tasks like human pose estimation and in a nutshell, which now defines the state-of-the-art in building reliable, real-world human sensing systems.

\label{sec:purify}
\begin{figure*}[t]
    \centering
    \includegraphics[width=\textwidth]{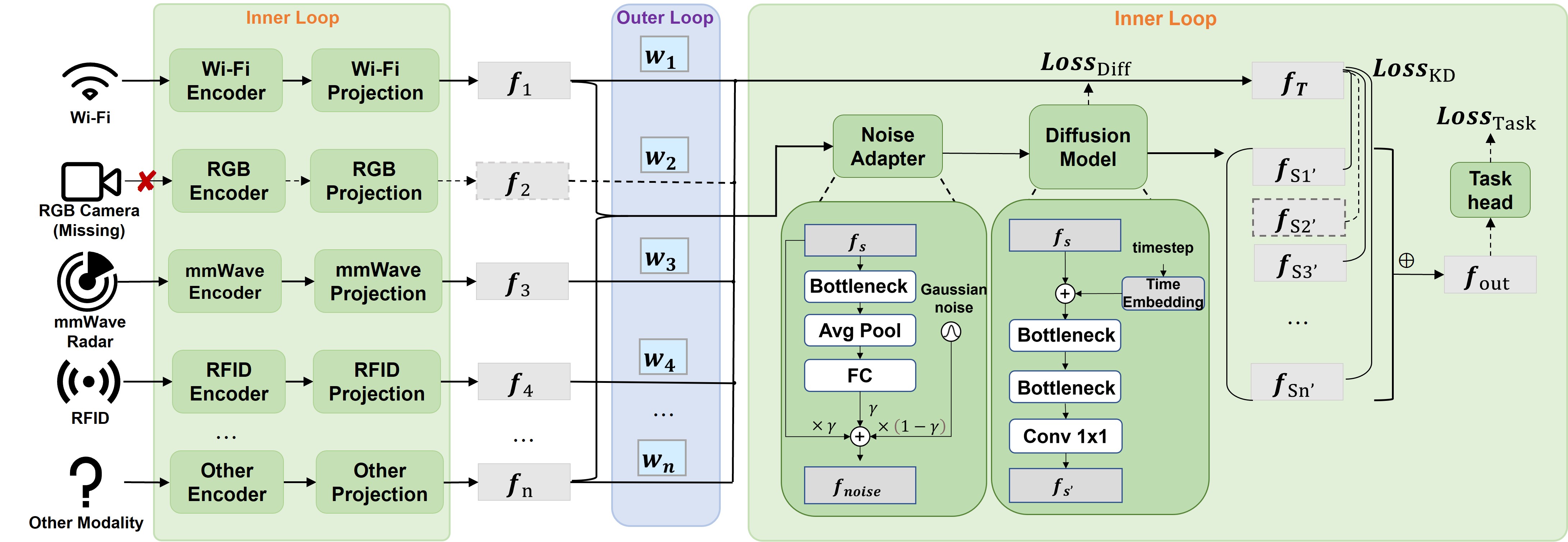}
    \caption{
    The overall architecture of our proposed \sysName{} framework, built on a Purify-then-Align paradigm. The model is trained in a nested loop: 
    (a) \textbf{The Purify Stage}: The Purify Stage optimizes the meta-learning weights $\mathbf{w}$ to mitigate the Contamination Effect. 
    (b) \textbf{The Align Stage}: The Align Stage uses these weights $\mathbf{w}$ to construct a ``clean teacher" $f_T$ and then uses it in a diffusion-based process ($\mathcal{L}_{\text{DiffKD}}$) to refine each ``student" feature $f_S$, thereby bridging the Representation Gap.
    }
    \label{fig:framework}
\end{figure*}

\subsection{Multimodal Learning under Modality Missing}\label{sec:mlmm}
Although multimodal human sensing offers more comprehensive and accurate measurements, in real-world scenarios, some modalities are often unavailable due to sensor instability or hardware deployment costs. Therefore, developing a framework that remains robust under modality-missing conditions has become a critical research direction in this field. Broadly, solutions for this problem follow two strategies. 
The first is modality generation, which synthesizes raw missing data using methods from simple imputation~\citep{parthasarathy2020training,cheng2024novel,yang2024incomplete} to generative models like GANs~\citep{maliakel2024fligan} and diffusion models~\citep{wang2023incomplete}, but they suffer from training instability, high computational costs, and a tendency to hallucinate inaccurate details.
The second strategy learns collaborative representation relationships in modalities~\citep{xu2024leveraging, wang2020icmsc, liu2021incomplete} or directly imputes the representation of missing modalities~\citep{li2025generating,yan2025federated,hoffman2016learning, he2022masked}, but it is often plagued by the difficulty of preserving unique and critical information from highly heterogeneous modalities.
 % is a prominent method

In human sensing, these two strategies are usually adapted and complemented by other techniques. For instance, These works~\citep{xue2022modality,kong2019mmact,wang2020incomplete,kwak2025cross} use knowledge distillation, where transferring knowledge from one powerful multi-modal or many single-modal teacher to a student. This enables the student to either infer the representation of a missing modality or enrich its own features, thus compensating for the incomplete input.
However, as demonstrated in \citep{bruce2021multimodal, xue2022modality}, this can require large pre-trained teachers and may be limited to specific tasks and attempts to bridge the representation gap are severely compromised by the Contamination Effect. In response to this, X-Fi~\citep{chen2025xfi} which builds a modality-invariant model to learn a compressed representation, 
 while it focuses on the quality of the fusion representation itself without addressing how this fusion result can empower the single-modality models to better infer missing representations. 
 Furthermore, X-Fi can be highly sensitive to the selection probability of each modality during training, leading to unstable performance.

\section{Method}\label{sec:method}

In this section, we present \sysName, a novel framework built upon the ``Purify-then-Align" paradigm to address the pervasive challenge of missing modalities \citep{ma2021smil, wang2020incomplete}. As mentioned in Sec. \ref{sec:introduction}, our approach aims to resolve the causal link between two fundamental challenges: the Representation Gap and the Contamination Effect, structured in two sequential and synergistic stages to resolve this dependency. Unlike existing methods that rely on fixed or manually tuned modality fusion strategies ~\citep{marcard2016human, das2021mmhar, chen2025xfi}, our framework first enters the ``Purify" stage. This stage directly tackles the Contamination Effect by introducing a meta-learning-driven Modality Weighting mechanism ~\citep{wang2024metakd}, which actively learns to down-weight noisy modalities, thereby ensuring a clean teacher representation. Subsequently, and only after this knowledge source is purified, the framework proceeds to the ``Align" stage. Here, we design a diffusion-based feature-level knowledge distillation \citep{ho2020ddpm} where this clean, many-to-one consensus teacher is leveraged to refine and align the features from each student modality. Through this principled design, \sysName enables the model to maintain robust performance even when modalities are unavailable.

\subsection{Purifying the Teacher Representation via Meta-Learning}\label{sec:purifying}

As illustrated in Figure~\ref{fig:framework}, our pipeline's ``Purify" stage is designed to learn a set of dynamic importance weights $\mathbf{w}$ that mitigate the Contamination Effect, thereby ensuring a clean teacher representation. To achieve this, the pipeline first simulates real-world challenges by randomly dropping each input modality with a uniform probability during training.
\fei{This strategy eliminates the need for the complex and sensitive trial-and-error tuning of per-modality dropout rates, as required by X-Fi \citep{chen2025xfi}. Manual tuning of these hyperparameters becomes particularly difficult as the number of modalities increases, a limitation our method effectively circumvents by avoiding any hard-coded dropout assignments.}
The core of this stage is a nested optimization strategy based on meta-learning \citep{ma2021smil}, which decouples the optimization of the whole sensing model parameters, $\Theta$, from the meta-parameters $\mathbf{w}$ that control the relative importance of each modality.

The optimization process consists of two nested loops. The inner loop aims to optimize the whole model parameters $\Theta$ (including encoders, task heads, and the alignment module detailed in Sec.~\ref{sec:aligning}) on the training set $\mathcal{D}_{\text{train}}$. During this loop, the modality weights $\mathbf{w}$ are treated as fixed hyperparameters. The objective is to minimize a composite loss function:
\begin{equation}
    \mathcal{L}_{\text{inner}} = \mathcal{L}_{\text{task}}(\Theta) + \lambda \mathcal{L}_{\text{DiffKD}}(\Theta),
    \label{eq:inner_loss}
\end{equation}
where $\mathcal{L}_{\text{DiffKD}}$ is our proposed feature alignment loss (detailed in Sec.~\ref{sec:aligning}) and $\lambda$ is a balancing coefficient. After the inner-loop update yields optimized parameters $\Theta^*$, the outer loop aims to optimize the meta-parameters $\mathbf{w}$. This is achieved by evaluating the performance of the inner-loop-trained model $\Theta^*(\mathbf{w})$ on a disjoint validation set, $\mathcal{D}_{\text{val}}$. The outer-loop loss is:
\begin{equation}
\mathcal{L}_{\text{outer}} = \mathcal{L}_{\text{task}}(\Theta^*(\mathbf{w})).
\label{eq:outer_loss}
\end{equation}
The gradient $\nabla_{\mathbf{w}} \mathcal{L}_{\text{outer}}$ is then computed to update $\mathbf{w}$. Formally, this entire nested optimization problem, which defines our purification process, can be expressed as:

\begin{align}
  \mathbf{w}^* &= \operatorname*{arg\,min}_{\mathbf{w}} \sum_{(\mathcal{M}_{\text{val}}, y) \in \mathcal{D}_{\text{val}}} \mathcal{L}_{\text{task}}\left(h^{*}\left(\sum_{i \in \mathcal{M}_{\text{val}}} w_i f'^{*}_{i}\right), y\right) \notag \\
  & \quad \text{s.t.} \quad \Theta^* = \operatorname*{arg\,min}_{\Theta} \sum_{(\mathcal{M}_{\text{train}}, y) \in \mathcal{D}_{\text{train}}} \notag \\
  & \quad \left[ \mathcal{L}_{\text{task}}\left(h\left(\sum_{i \in \mathcal{M}_{\text{train}}} w_i f_i\right), y\right) + \lambda \mathcal{L}_{\text{DiffKD}}(\Theta) \right]
\end{align}
where $h$ and $h^{*}$ represent the task-specific head before and after the inner-loop update, respectively. $f_i$ and $f'^{*}_{i}$ are the original and distilled features from modality $i$ (distillation is detailed in Sec.~\ref{sec:aligning}). To ensure stability and interpretability, the weights $\mathbf{w}$ are normalized via a Softmax function after each outer-loop update:
\begin{equation}
    \mathbf{w}_i \leftarrow \frac{\exp(\mathbf{w}_i)}{\sum_{j=1}^{N} \exp(\mathbf{w}_j)}, \quad i=1,\dots,N.
    \label{eq:softmax}
\end{equation}

This meta-learning framework allows the model to jointly learn task-specific representations and modality importance in a principled, adaptive manner. This process achieves the central goal of this stage: reducing the Contamination Effect of low-quality modalities on the fused representation that will serve as the clean teacher in the subsequent ``Align" stage.

\begin{algorithm}[htbp]
\caption{The \sysName{} Algorithm: Purify-then-Align}
\begin{algorithmic}[1]
\REQUIRE $p(T)$: distribution over tasks
\REQUIRE $\Theta$: initial model parameters 
\REQUIRE $\mathcal{D}_{\text{train}}$: training dataset
\REQUIRE $\mathcal{D}_{\text{val}}$: validation dataset
\REQUIRE $\mathbf{w}$: meta-parameters

\STATE \textbf{Initialize} $f_T = \text{construct\_teacher()}$
\STATE \textbf{Initialize} $f_S = \text{initialize\_student()}$
\WHILE{not converged}
    \STATE Sample batch of tasks $T_i \sim p(T)$
    \FOR{all $T_i$}
        \STATE $z_T = \text{projection}(f_T)$
        \STATE $z_S = \text{projection}(f_S)$
        \STATE $noisy\_z_T = \text{adapt}(z_T)$
        \STATE $predicted\_noise = \text{predict\_noise}(noisy\_z_T)$
        \STATE $\mathcal{L}_{\text{Diff}} =
        \text{compute\_loss}(noisy\_z_T,$
        \STATE \hspace{1.5em} $predicted\_noise)$
        \STATE $\hat{z}_S = \text{refine\_student\_feature}(z_S)$
        \STATE $\mathcal{L}_{\text{KD}} = \text{compute\_KD\_loss}(\hat{z}_S, z_T)$
        \STATE $\mathcal{L}_{\text{inner}} = \mathcal{L}_{\text{task}} + \lambda \mathcal{L}_{\text{Diff}}$
        \STATE $\text{update}(\Theta, \mathcal{L}_{\text{inner}})$
    \ENDFOR
    
    \STATE $\mathcal{L}_{\text{outer}} = \mathcal{L}_{\text{task}}(\Theta^*(\mathbf{w}))$
    \STATE $\mathbf{w}^* = \text{update\_meta\_parameters}(\mathbf{w}, \mathcal{L}_{\text{outer}})$
\ENDWHILE
\STATE \textbf{Return} Optimized model parameters $\Theta^*$ and $\mathbf{w}^*$
\end{algorithmic}
\label{alg:Purify-then-Align}
\end{algorithm}

\subsection{Aligning Student Modalities with Knowledge Diffusion}
\label{sec:aligning} 

The previous section detailed the ``Purify" stage, a meta-learning process designed to find the optimal modality weights $\mathbf{w}^*$. Now, in this ``Align" stage, we formally describe the core distillation mechanism that drives this optimization. This mechanism, introduced as $\mathcal{L}_{\text{DiffKD}}(\Theta)$ in Equation~\ref{eq:inner_loss}, is the component responsible for bridging the Representation Gap. We design this alignment process as a diffusion-based feature alignment mechanism that integrates knowledge distillation with diffusion models \citep{ho2020ddpm, hinton2015distilling, NEURIPS2023_cdddf13f}. As shown in Figure~\ref{fig:framework}, this is a ``many-to-one" paradigm, where the ``teacher" feature, derived from a consensus of all available modalities, is used to guide the ``student" feature alignment from each modality. 

Specifically, the teacher feature, $f_T$, is constructed to represent the optimal fusion of all available modalities that serves as the distillation target. It is formally defined as the weighted sum of the features from all available modalities in the set $\mathcal{M}_{\text{all}}$:
\begin{equation}
    f_T = \sum_{i \in \mathcal{M}_{\text{all}}} \mathbf{w}_i f_i,
    \label{eq:teacher_feat}
\end{equation}
where $f_i$ is the feature of modality $i$ after passing through its specific encoder and a projection layer, and $\mathbf{w}_i$ is its corresponding meta-learned importance weight. This construction ensures that $f_T$ is a comprehensive representation. The student feature, $f_S$, is defined as the encoded and projected feature of any single available modality. The goal of our alignment process is to transform the information-deprived $f_S$ from a single modality into a feature that structurally aligns with the rich teacher feature $f_T$. 

To achieve this alignment, we first map the features into a compressed latent space (a lightweight design choice detailed in Sec.~\ref{sec:lightweight_adapter}). As shown in Figure~\ref{fig:framework}, the teacher feature $f_T$ and student feature $f_S$ are passed through projection layers to create latent representations, $z_T$ and $z_S$, respectively.

We then employ a diffusion-based framework, conceptualizing $z_S$ as a ``noisy'' or incomplete version of $z_T$. The core of this process is a noise prediction network, $\Phi_\phi$, trained exclusively on the ``clean'' latent teacher features $z_T$ to learn their underlying data distribution. The training objective is the standard diffusion loss, $\mathcal{L}_{\text{Diff}}$:\begin{equation}
\label{eq:diff_loss_single}
\begin{split}
\mathcal{L}_{\text{Diff}} &= \mathbb{E}_{t, z_T, \epsilon_t} \left[ ||\epsilon_t - \Phi_{\phi}(z_t, t)||_2^2 \right], \\
\text{where } z_t &= \sqrt{\bar{\alpha}_t} z_T + \sqrt{1-\bar{\alpha}_t} \epsilon_t.
\end{split}
\end{equation}
Here, $t$ is the diffusion timestep, $\epsilon_t \sim \mathcal{N}(0, \mathbf{I})$ is the sampled Gaussian noise, and $\bar{\alpha}_t$ is a parameter derived from the predefined noise schedule. How this alignment process is made lightweight and adaptive to the student's noise level will be detailed in Sec.~\ref{sec:lightweight_adapter}.

This entire alignment mechanism is encapsulated in the total feature alignment loss $\mathcal{L}_{\text{DiffKD}}$, which was introduced in Equation~\ref{eq:inner_loss}. It is defined as the sum of the diffusion loss $\mathcal{L}_{\text{Diff}}$ and a separate knowledge distillation loss $\mathcal{L}_{\text{KD}}$:
\begin{equation}
    \mathcal{L}_{\text{DiffKD}} = \mathcal{L}_{\text{Diff}} + \mathcal{L}_{\text{KD}},
    \label{eq:total_loss} 
\end{equation}
where $\mathcal{L}_{\text{KD}}$ is the core distillation loss, defined as a distance (e.g., MSE loss) between the refined student latent feature $\hat{z}_S$ (generated in Sec.~\ref{sec:lightweight_adapter}) and the latent teacher feature $z_T$.
This design ensures that even when only partial modalities are available, the student feature is refined to approximate a complete multimodal representation, improving robustness and modality-invariance.

\subsection{Lightweight Feature Alignment and Adaptive Denoising}
\label{sec:lightweight_adapter}
As introduced in Sec.~\ref{sec:aligning}, a key challenge is to make the alignment process both computationally efficient and adaptive to the student's unknown noise level. Our framework addresses both issues through our lightweight and adaptive design. To achieve computational efficiency, we operate in a compressed latent space. Each modality's feature $f_i$ is first passed through its respective projection layer (a 1x1 convolution), mapping the features $f_T$ and $f_S$ into lower-dimensional latent representations. The diffusion model $\Phi_\phi$ (from Equation~\ref{eq:diff_loss_single}), which operates exclusively in this latent space, is also lightweight. As shown in Figure~\ref{fig:framework}, it is composed of stacked Bottleneck blocks and a final convolution 1x1 layer. This combined latent-space approach is validated in \citep{NEURIPS2023_cdddf13f} for efficient diffusion.

Our framework must also solve the adaptivity problem. Since the student's latent feature $z_S$
is an incomplete representation from $f_S$, its noise level relative to $z_T$
differs for each input sample. This sample-dependent relationship fundamentally requires a dynamic adjustment of the denoising process, as a fixed diffusion timestep $t$ cannot represent this one-to-many mapping.
To solve this, we introduce the Noise Adapter. As detailed in Figure~\ref{fig:framework}, this module performs adaptive noise matching by mapping $z_S$ to a well-defined starting point for the reverse process. Architecturally, it is a small auxiliary network (composed of 1 Bottleneck block, a global Average Pooling layer, and a Fully-Connected layer) that predicts a scalar fusion coefficient, $\gamma \in [0, 1]$. This coefficient is then used to optimally blend the student's latent feature $z_S$ with pure Gaussian noise $\epsilon_T$ (sampled at the maximum timestep $T$):
\begin{equation}
z_{TS} = \gamma z_S + (1-\gamma) \epsilon_T, \quad \text{where } \epsilon_T \sim \mathcal{N}(0, \mathbf{I}).
\label{eq:noise_adapter}
\end{equation}
The resulting feature, $z_{TS}$, serves as the high-quality starting point for the reverse denoising process. Finally, the reverse process is also executed efficiently by using the Denoising Diffusion Implicit Models (DDIM) sampling strategy \citep{song2021ddim}. DDIM replaces the full, stochastic $T$-step (e.g., $T=1000$) denoising chain with a short, deterministic one. We define a small number of inference steps $N_{steps}=5$, which we found to be an effective trade-off between alignment performance and inference speed, creating a sampling interval $\Delta = T / N_{steps}$.  The model then efficiently reverses the process in 5 deterministic steps ($z_{T} \rightarrow z_{T-\Delta} \rightarrow z_{T-2\Delta} \rightarrow \dots \rightarrow z_{0}$), yielding the final aligned and refined student latent feature, $\hat{z}_S$. 

\fei{The pseudocode of the PTA algorithm process is presented in Algorithm~\ref{alg:Purify-then-Align}.}
Overall, the above lightweight and adaptive approach ensures the feature alignment is both efficient and effective.

\section{Evaluation}\label{sec:evaluation}

\fei{
We evaluate \sysName on two large-scale multimodal human sensing benchmarks: MM-Fi~\cite{yang2023mm} and XRF55~\cite{wang2024xrf55}. These datasets exhibit significant heterogeneity in their modalities (e.g., depth, wireless signals, LiDAR), leading to a pronounced Representation Gap. Furthermore, they contain real-world noise and low-quality modalities, making them ideal for validating our framework's robustness against the Contamination Effect. We follow the standard data splits and preprocessing procedures established in the state-of-the-art work, X-Fi~\cite{chen2025xfi}, to ensure a fair comparison.}

\subsection{Datasets}\label{sec:datasets}

\textbf{MM-Fi}~\citep{yang2023mm} is a large-scale, multimodal dataset designed for non-intrusive human pose estimation. It provides more than 320,000 synchronized frames collected from 40 human subjects performing 27 distinct actions. The dataset includes five sensing modalities, and three of them are publicly available: depth image, LiDAR point cloud, and Wi-Fi channel state information (CSI) data. This rich combination of high and low-resolution sensors makes MM-Fi an ideal benchmark for evaluating the robustness of multimodal human sensing frameworks, especially under challenging missing modality conditions. We used the official data splits and evaluation protocols for our experiments. 

\textbf{XRF55}~\citep{wang2024xrf55} is another large-scale, multimodal radio frequency (RF) dataset designed for human indoor action recognition. It contains 42,900 RF samples, totaling over 59 hours of data, collected from 39 human subjects performing 55 distinct actions, covering categories such as fitness and human-computer interactions. The dataset's key feature is its use of diverse, commercial off-the-shelf RF sensing modalities, including one mmWave radar (60-64GHz) providing Range-Doppler \& Range-Angle Heatmaps (R), 9 Wi-Fi links capturing Wi-Fi CSI at 5.64GHz (W), and 23 RFID tags providing phase series data at 922.38 MHz. in our experiments, we follow the original data split setting for XRF55 as outlined in its paper.

\subsection{Implementation Details}\label{sec:implementation-details}
\textbf{On MM-Fi dataset.} For the human pose estimation (HPE) task on the MM-Fi dataset, the modality encoders for the Depth, LiDAR, and Wi-Fi modalities are based on pre-trained and frozen models in X-Fi~\citep{chen2025xfi}. We standardize the feature representation of each modality to a dimension of 512 ($d_f=512$). The noise prediction network within the knowledge diffusion module is implemented as a lightweight residual network, composed of two consecutive bottleneck blocks followed by a final 1D convolutional layer; the feature alignment itself is performed using a DDIM sampling strategy with 5 inference steps. Our \sysName framework is optimized using two separate Adam optimizers: one for the main model parameters with a learning rate of $5\times10^{-4}$, and one for the meta-learning weights with a learning rate of $1\times10^{-2}$. The balancing coefficient $\lambda$ for the knowledge distillation loss is set to 0.1. All experiments are conducted on an NVIDIA RTX 3090 GPU with a batch size of 16. We use the Mean Per Joint Position Error (MPJPE) and Procrustes-Aligned MPJPE (PA-MPJPE) as evaluation metrics, where lower values indicate better performance.

\textbf{On XRF55 dataset.} For the human action recognition (HAR) task on the XRF55 dataset, the modality encoders for all RF modalities (mmWave Radar, Wi-Fi CSI, and RFID) are based on a pre-trained and frozen ResNet-18 architecture. We standardize the feature representation of each modality to 32 features ($n_f=32$), each with a dimension of 512 ($d_f=512$). The noise prediction network and alignment strategy are identical to those used for the MM-Fi dataset. Similarly, our \sysName framework is optimized using the Adam optimizer with a learning rate of $2 \times 10^{-4}$ and $\lambda=0.1$. All experiments are conducted on an NVIDIA RTX 3090 GPU with a batch size of 32. We use classification accuracy as the evaluation metric, where higher values indicate better performance.

\subsection{Human Pose Estimation Results on the MM-Fi Dataset}\label{sec:result-mmfi}

 We compare \sysName against three key methods: a feature-fusion baseline where modality-specific features are concatenated and processed by an MLP fusion head \citep{das2021mmhar}~(Base.1), a decision-level baseline that averages the final prediction outputs from each modality \citep{yang2023mm}~(Base.2), and the state-of-the-art modality-invariant model, X-Fi \citep{chen2025xfi}.

\begin{table*}[t]
% \small
\centering
\small
\setlength{\tabcolsep}{3pt}
\caption{MPJPE~(mm) and PA-MPJPE~(mm) on MM-Fi~\citep{yang2023mm} dataset.}
\label{tab:mmfi-results}
% \resizebox{\textwidth}{!}{%
\begin{tabular}{l|llccl|llccl}
\toprule
Modality & Base.1 & Base.2 & X-Fi & Ours & Improvement$\uparrow$ &
Base.1 & Base.2 & X-Fi & Ours & Improvement$\uparrow$ \\ 
\midrule
& \multicolumn{5}{c}{MPJPE $\downarrow$} &
  \multicolumn{5}{c}{PA-MPJPE~$\downarrow$} \\ 
\midrule
D & 102.4 & 102.4 & 96.40 & \textbf{84.81} & +12.0\% &
52.7 & 52.7 & \textbf{48.76} & 50.72 & -4.0\% \\

L & 161.5 & 161.5 & 130.06 & \textbf{68.30} & +47.5\% &
103.5 & 103.5 & 89.83 & \textbf{44.44} & +50.5\% \\

W & 227.1 & 227.1 & 210.12 & \textbf{182.18} & +13.3\% &
108.0 & 108.0 & \textbf{103.91} & 114.63 & -10.3\% \\

D+L & 111.7 & 108.0 & 89.41 & \textbf{64.68} & +27.7\% &
68.8 & 55.8 & 47.97 & \textbf{42.63} & +11.1\% \\

D+W & 141.7 & 155.5 & \textbf{95.27} & 95.96 & -0.7\% &
71.4 & 81.5 & \textbf{47.71} & 59.77 & -25.3\% \\

L+W & 167.1 & 206.2 & 111.15 & \textbf{74.74} & +32.8\% &
100.7 & 109.3 & 75.11 & \textbf{49.49} & +34.1\% \\

D+L+W & 130.7 & 154.6 & 87.59 & \textbf{68.86} & +21.4\% &
78.1 & 84.1 & 47.52 & \textbf{45.47} & +4.3\% \\ 
\midrule
Best Count & 0 & 0 & 1 & \textbf{6} & - &
0 & 0 & 3 & \textbf{4} & - \\ 
\midrule
\end{tabular}%
% }
\end{table*}

\textbf{Finding 1: The Purify-then-Align mechanism greatly empowers the representation of a single modality.}

As shown in Table~\ref{tab:mmfi-results}, our proposed method significantly increases the performance of each modality in isolation compared to both the baseline and X-Fi. For example, the individual performance of the Depth, LiDAR and Wi-Fi CSI modalities saw remarkable improvements of 12.0\%, 47.5\% and 12.3\%, respectively, over the X-Fi method. This substantial gain validates the efficacy of our core mechanism: \sysName fundamentally enhances the robustness of each single-modality feature by first purifying the information-rich teacher features (which are extracted from the consensus of all modalities) and subsequently aligning the student (single-modality) features. In contrast, methods like X-Fi primarily focus on the collaborative adaptation of features purely for fusion purposes, which fails to significantly elevate the intrinsic quality of the individual single-modality representations. By compelling each single-modality student feature to be refined by the higher-quality, purified teacher feature, our framework substantially elevates the inherent capability of every feature extractor.

\textbf{Finding 2: Efficacy of the ``Purify'' Stage in Contamination Mitigation}

As shown in Table~\ref{tab:mmfi-results}, the Wi-Fi modality, with a standalone MPJPE of 182.18, performs significantly worse than Depth (84.81) and LiDAR (68.30), establishing it as a low-quality modality for this task. The introduction of such a modality risks contaminating the fused representation and degrading overall performance. Our results demonstrate that the meta-learning weighting mechanism effectively mitigates this risk. For instance, when fusing LiDAR with Wi-Fi (L+W), our method achieves an MPJPE of 74.74, showing only a minor degradation from LiDAR's standalone performance of 68.30. This indicates that \sysName successfully identified Wi-Fi as a low-contributor and down-weighted its influence, thus preventing significant contamination. This ability confirms that the ``Purify'' stage successfully alleviates the Contamination Effect, laying a robust foundation for the subsequent ``Align'' stage.

Interestingly, X-Fi's performance for L+W (111.15) is better than its standalone LiDAR performance (130.06). X-Fi sacrifices the maximum performance of its base single-modality models (as evidenced by its LiDAR MPJPE of 130.06 vs. our 68.30) in exchange for marginal fusion gains. Our results suggest this is an inefficient trade-off, as our method's superior single-modality performance provides a much stronger foundation, leading to a more favorable and robust overall performance.

\textbf{Discussion on MPJPE vs. PA-MPJPE.}

An interesting trade-off is observed between the absolute error (MPJPE) and structure-aligned error (PA-MPJPE). Our method's superior MPJPE performance stems from the knowledge diffusion process effectively extracting and enhancing global positioning information from noisy signals like CSI, allowing the model to better determine where a person is. However, these same signals lack the local details required for precise limb placement. Consequently, after the Procrustes Alignment in PA-MPJPE nullifies the global advantage, the evaluation focuses on the skeleton's internal structure. Lacking fine-grained local information, our model may converge on a structurally plausible but incorrect ``average pose'', causing a slight increase in the PA-MPJPE.  Future hybrid approaches that integrate biomechanical constraints or pose priors could potentially help.

\subsection{Human Action Recogntion Results on the XRF55 Dataset}\label{sec:result-xrf55}

 We compare \sysName with two methods: a mutual learning method implemented XRF55 \citep{wang2024xrf55}~(Base.) and the state-of-the-art X-Fi \citep{chen2025xfi}.

\textbf{Echoing Finding 1: The Purify-then-Align mechanism greatly empowers the representation of the single modality.}

As shown in Table~\ref{tab:har_xrf55}, \sysName demonstrates a significant performance increase for the single modality compared to X-Fi. Specifically, the mmWave Radar (R), Wi-Fi (W), and RFID (RF) modalities achieve accuracy improvements of +6.13\%, +26.64\%, and +12.54\%, respectively. 
This significant uptick in performance is a direct result of our knowledge diffusion mechanism, which echoes the core principle of MM-Fi. Our system trains the feature extractor of each modality to align its representation with a comprehensive teacher feature, which is synthesized from the consensus of all modalities. This process effectively allows low-quality modalities, such as Wi-Fi, to learn from the more robust and detailed patterns captured by other modalities. Moreover, X-Fi requires carefully tuned, hard-coded dropout rates and is highly sensitive to these specific values. For instance, when the dropout rates for Radar, Wi-Fi, and RFID were set to (0.5, 0.5, 0.8), (0.5, 0.7, 0.7), and (0.5, 0.9, 0.6) respectively, the accuracy for the Wi-Fi modality varied significantly at 29.1\%, 37.4\%, and 55.7\%. In contrast, our proposed \sysName utilizes uniform modality dropout, which is not only simpler but also delivers more stable performance.

\begin{table}[t]
\small
    \centering
    \caption{Accuracy (\%) on the XRF55 dataset.}
    \label{tab:har_xrf55}
    \begin{tabular}{lrrrr}
    \toprule
    Modality & Base. & X-Fi & Ours & Improvement $\uparrow$ \\
    \midrule
    R & 82.1 & 83.9 & \textbf{90.03} & +6.13 \\
    W & 77.8 & 55.7 & \textbf{82.34} & +26.64 \\
    RF & 42.2 & 42.5 & \textbf{55.04} & +12.54 \\
    R+W & 86.8 & 88.2 & \textbf{95.23} & +7.03 \\
    R+RF & 71.4 & 86.5 & \textbf{92.15} & +5.65 \\
    W+RF & 55.6 & 58.1 & \textbf{83.14} & +25.04 \\
    R+W+RF & 70.6 & 89.8 & \textbf{95.87} & +6.07 \\ \midrule
    Best Count & 0 & 0 & \textbf{7} & - \\
    \bottomrule
    \end{tabular}
\end{table}

\textbf{Finding 2: Meta-Learning Mitigates Contamination and Can Yield Positive Fusion Gains.}

In the MM-Fi results, we observed that a low-quality modality could contaminate a high-quality one, leading to fusion performance that was worse than the high-quality modality alone. However, our findings on XRF55 are more optimistic and show this is not always the case. As shown in Table~\ref{tab:har_xrf55}, RFID is the worst modality with a standalone accuracy of only 55.04\%. Yet, when fused with Radar, it improves the accuracy from 90.03\% to 92.15\%. Similarly, fusing it with Wi-Fi improves accuracy from 82.34\% to 83.14\%. Furthermore, when all modalities are present, our performance reaches 95.87\%, significantly outperforming the baseline (70.6\%) and X-Fi (89.8\%). This demonstrates that under certain conditions, our framework effectively resists the Contamination Effect and leverages low-quality modalities to achieve performance gains.

We hypothesize this occurs because the modalities in XRF55 (mmWave, Wi-Fi, RFID) are all RF-based signals. The representation gap between them is smaller than the gap between the heterogeneous modalities in MM-Fi (Wi-Fi, Depth, LiDAR), allowing \sysName to learn more effective alignment and fusion strategies. An open question remains regarding the threshold at which a modality becomes so poor that it actively contaminates others. While RFID is the worst performer here, it may not be weak enough to cause significant harm, unlike the CSI data in MM-Fi. This aligns with findings in NLP where dirty data can degrade performance \citep{wallace2021concealed}, but it remains a less-explored and intriguing area for future research in the human sensing domain.

\subsection{Ablation Study}\label{sec:ablation-study}

To validate the effectiveness of core components within proposed in \sysName, we systematically removed the Diffusion-based Alignment module and the meta-learning weighting module to independently evaluate their respective contributions on both the MM-Fi and XRF55 datasets. The results are presented in Table~\ref{tab:albation-mmfi} and Table~\ref{tab:ablation_xrf55}, respectively.

\begin{table}[htbp]
\small
\centering
\begin{minipage}{.48\textwidth}
  \centering
  \small
  \captionof{table}{Ablation study of our \sysName model on the MM-Fi HPE task, (MPJPE~(mm)).}
  \label{tab:albation-mmfi}
  \begin{tabular}{lccc}
    \toprule
    Modality & Full & w/o Diff. & w/o Meta.\\
    \midrule
    D & \textbf{84.81} & 89.66 & 157.98 \\
    L & \textbf{68.30} & 76.27 & 183.04 \\
    W & \textbf{182.18} & 187.92 & 236.99 \\
    D+L & \textbf{64.68} & 78.12 & 148.65 \\
    D+W & 95.96 & \textbf{90.14} & 179.77 \\
    L+W & 74.74 & \textbf{74.23} & 189.24 \\
    D+L+W & \textbf{68.86} & 76.79 & 160.34 \\ \midrule
    Best count & \textbf{5} & 2 & 0\\
    \bottomrule
  \end{tabular}
\end{minipage}%
\hfill 
\begin{minipage}{.48\textwidth}
  \centering
  \small
  \captionof{table}{Ablation study of our \sysName model on the XRF55 HAR task, (Accuracy\%).}
  \label{tab:ablation_xrf55}
  \begin{tabular}{lccc}
    \toprule
    Modality & Full & w/o Diff. & w/o Meta. \\
    \midrule
    R   & \textbf{90.03} &  89.17 & 89.04 \\
    W   & \textbf{82.34} &  81.07 & 79.56\\
    RF  & \textbf{55.04} &  54.49 & 54.79 \\
    % \midrule
    R+W & 95.23 &  \textbf{95.40} & 93.61 \\
    R+RF & 92.15 &  91.60 & \textbf{92.87} \\
    W+RF & 83.14 &  \textbf{84.35} & 78.35 \\
    % \midrule
    R+W+RF & \textbf{95.87} &  95.59 & 95.07 \\ \midrule
    Best count & \textbf{4} & 2 & 1\\
    \bottomrule
  \end{tabular}
\end{minipage}
\end{table}

Table~\ref{tab:albation-mmfi} and Table~\ref{tab:ablation_xrf55} confirm that the full \sysName framework consistently achieves the best performance, securing the top result in 5 out of 7 modality combinations on the MM-Fi human pose estimation task and 4 out of 7 on the XRF55 human action recognition task. This demonstrates that the two modules are not merely independent components but are synergistically integrated. They work together effectively to enhance performance under missing modality conditions, forming a cohesive and powerful framework.

When without Diffusion-based Alignment module (``w/o Diff."), we observe a notable performance drop, particularly in single-modality scenarios (e.g., LiDAR MPJPE increases from 68.30 to 76.27). The effect is even more pronounced when the meta-learning weighting module is removed (``w/o Meta."). In this case, performance degrades catastrophically across nearly all fusion scenarios (e.g., D+L MPJPE skyrockets from 64.68 to 148.65), providing clear evidence of its success in mitigating the Contamination Effect by controlling the influence of disparate modalities. 

A more nuanced analysis of the ablation studies reveals an interesting edge case. While the Diffusion-based Alignment module is overwhelmingly beneficial, we observed a few specific scenarios where its removal paradoxically improved performance. As shown in Table~\ref{tab:albation-mmfi} (L+W) and Table~\ref{tab:ablation_xrf55} (W+RF), this phenomenon occurs when fusing two modalities where at least one is of significantly low quality (Wi-Fi or RFID). We hypothesize that this is due to the generative nature of the diffusion process. While the model's primary function is to align features and bridge the representation gap, the reverse sampling process can still introduce minor generative artifacts or amplify inherent noise, especially when the conditioning student feature is extremely noisy or information-poor.

\section{Conclusion}\label{sec:conclusion}

We introduced \sysName, a novel framework that addresses two causally linked challenges in robust multimodal human sensing: the Representation Gap and the Contamination Effect. We identify the Contamination Effect, where noisy modalities corrupt the teacher's representation,  as the key bottleneck preventing the bridging of the Representation Gap. \sysName introduces a synergistic ``Purify-then-Align" paradigm: a meta-learning weighting mechanism first purifies the teacher by suppressing low-quality modalities, followed by a diffusion-based distillation strategy that aligns each student modality using the refined teacher. Extensive experiments on large-scale MM-Fi and XRF55 datasets validate the framework's effectiveness. 

This work contributes a vital link to wireless sensing generalizability during model development~\cite{wang2026survey}, inspiring subsequent research in this evolving domain.

% The result is a set of powerful single-modality encoders, empowered with clean cross-modal knowledge, that remain highly robust even under severe modality-missing conditions.

\noindent \textbf{Acknowledgments:}
This work was supported by the National Natural Science Foundation of China under grant 62572383 and Fundamental Research Funds for the Central Universities.

%%%%%%%%% REFERENCES
{
    \small
    \bibliographystyle{ieeenat_fullname}
    \bibliography{main}
}

\end{document}